\documentclass[conference]{IEEEtran}
\IEEEoverridecommandlockouts
\usepackage{url}
\usepackage{microtype}
\usepackage{graphicx}
\usepackage{booktabs}
\usepackage{multirow} 
\usepackage[font={small}]{caption}
\usepackage{amsmath}
\usepackage{amssymb}
\usepackage{mathtools}
\usepackage{amsthm}

\usepackage{cite}
\usepackage{subcaption}
\usepackage{algorithm}
\usepackage{algorithmic}
\usepackage{amsthm}
\usepackage{mathrsfs}
\usepackage{lineno}
\usepackage{bm}
\usepackage{tikz,xcolor,hyperref}
\usepackage{algorithm, algorithmic}
\usepackage{lettrine}
\usepackage{orcidlink}

\usepackage{balance}

\newcommand{\B}{\mathbf}

\def\MyMethod{MD-Face} 
\def\MyNetwork{MD-Net}

\foreach \x in {A, ..., Z}{%
 \expandafter\xdef\csname
 orcid\x\endcsname{\noexpand\href{https://orcid.org/\csname orcidauthor\x\endcsname}{\noexpand\orcidicon}}
}

\title{\MyMethod: MoE-Enhanced Label-Free Disentangled Representation for Interactive Facial Attribute Editing}
\def\BibTeX{{\rm B\kern-.05em{\sc i\kern-.025em b}\kern-.08em
    T\kern-.1667em\lower.7ex\hbox{E}\kern-.125emX}}
\begin{document}

\author{Xuan Cui, Yunfei Zhao, Bo Liu, Wei Duan, Xingrong Fan}

\maketitle
\begin{abstract}
  GAN-based facial attribute editing is widely used in virtual avatars and social media but often suffers from attribute entanglement, where modifying one face attribute unintentionally alters others. While supervised disentangled representation learning can address this, it relies heavily on labeled data, incurring high annotation costs. To address these challenges, we propose \MyMethod, a label-free disentangled representation learning framework based on Mixture of Experts (MoE). \MyMethod\ utilizes a MoE backbone with a gating mechanism that dynamically allocates experts, enabling the model to learn semantic vectors with greater independence. To further enhance attribute entanglement, we introduce a geometry-aware loss, which aligns each semantic vector with its corresponding Semantic Boundary Vector (SBV) through a Jacobian-based pushforward method. Experiments with ProGAN and StyleGAN show that \MyMethod\ outperforms unsupervised baselines and competes with supervised ones. Compared to diffusion-based methods, it offers better image quality and lower inference latency, making it ideal for interactive editing.
\end{abstract}
  
  
  \section{Introduction}\label{sec:introduction}
  
  \IEEEPARstart{F}acial attribute editing plays a crucial role in various multimedia applications, such as virtual avatar customization~\cite{jiang2022nerffaceediting}, social media filters, digital content creation~\cite{Xu2022}, and privacy-preserving~\cite{he2024}. Its goal is to manipulate a specific facial attribute (e.g., age) in an image, while maintaining both the identity and structural integrity of the face. Existing methods can be broadly divided into two main approaches: diffusion models and generative adversarial networks (GANs). While diffusion models~\cite{dds2023,cds2024} are capable of generating high-quality images, they are often plagued by significant sampling-time delays and high computational demands, making them less ideal for real-time or large-scale applications. In contrast, GAN-based methods~\cite{Sun2024WEMGANWT, 2025Facial,9645200} provide faster inference, making them better suited for interactive and real-time multimedia editing. However, these methods frequently suffer from the issue of cross-attribute entanglement: modifying a target attribute (e.g., age) may unintentionally affect non-target attributes (e.g., smile), which hampers fine-grained control. Recent advances in disentangled representation learning (DRL)~\cite{DRL2024} have successfully disentangled different facial attributes by learning attribute-specific semantic vectors in the GAN's latent space, enabling more precise attribute control.\par
  Despite these advancements, many DRL methods, especially supervised ones~\cite{EnjoyGAN2021,MaskFaceGAN2023,MDSE2023}, rely heavily on labeled datasets to improve disentanglement, incurring significant annotation costs. This motivates the development of unsupervised (label-free) DRL techniques, which discover semantic vectors without the need for labeled data, thereby reducing annotation costs while preserving strong disentanglement. Based on this motivation, we propose \MyMethod, a label-free DRL framework based on Mixture of Experts (MoE) operating in the GAN's latent space. The key contributions of this work are as follows:\par
  $\bullet$ We introduce a MoE-based Disentangling Network (\MyNetwork), where experts are dynamically allocated through a gating mechanism, with each expert focusing on learning different semantic vectors to achieve the disentangling of facial attributes.\par
  $\bullet$ We propose a geometry-aware loss that aligns each learned semantic vector with its corresponding Semantic Boundary Vector (SBV) through geometry-aware alignment. This loss not only improves the consistency between semantic vectors and their target attributes but also effectively reduces cross-attribute interference, enhancing attribute disentanglement. \par
  $\bullet$ We demonstrate the superior performance of \MyMethod\ across multiple pre-trained GAN models, surpassing existing unsupervised (label-free) methods, through extensive experimental comparisons. It also outperforms supervised models and diffusion-based models, particularly excelling in image quality metrics (e.g., FID, LPIPS) and attribute disentanglement metrics (e.g., AA, IDS).
  
  \section{Related work}\label{sec:related_work}
  \subsection{Supervised DRL for GAN-based facial attribute editing}
  Supervised DRL methods in the GAN's latent space typically utilizes labeled datasets or signals from pre-trained attribute classifiers to learn semantic vectors that control facial attributes. For example, MDSE~\cite{MDSE2023} proposed an orthogonal latent decomposition method guided by attribute classification, aimed at reducing face attributes entanglement. EnjoyGAN~\cite{EnjoyGAN2021} introduced a linear attribute regressor and used prediction scores as regularizers to promote the disentanglement of semantic directions. InterFaceGAN~\cite{InterFaceGAN2020} trained support vector machine (SVM) decision boundaries on labeled attributes, enabling the identification of separable directions and facilitating fine-grained, continuous control. AttGAN~\cite{AttGAN2019} combined attribute classification loss with pixel-level reconstruction to modify specific attributes while preserving non-target features. The StarGAN family~\cite{StarGAN2017,StarGAN2019} addressed multi-attribute and multi-domain translation by sharing parameters and domain-specific normalization, allowing scalable control within a unified model. Despite achieving promising results, these methods rely heavily on the availability and quality of labels, which may lead to high annotation costs, limited attribute coverage, and dataset biases, ultimately limiting generalization and scalability.
  
  \subsection{Unsupervised (Label-Free) DRL for facial attribute editing}
  Unsupervised (Label-Free) DRL methods eliminate the need for explicit annotations by utilizing the inherent structural characteristics of the latent or image spaces. For latent space direction discovery, techniques like those in GANSpace~\cite{GANSpace2020}, Escape~\cite{ICLR2022Escape}, SeFa~\cite{SEFA2021}, and AdaTrans~\cite{AdaTrans2023} leveraged methods such as principal component analysis, linear semantic decomposition, eigen decomposition of generator weights, and adaptive nonlinear transformations to identify meaningful directions for editing. In addition, StyleFlow~\cite{StyleFlow2021} and SDFlow~\cite{SDFlow2024ICASSP} adopted normalizing flows to capture nonlinear relationships in the latent space, often with weak supervision via attribute conditioning. In terms of image space translation, methods like CycleGAN~\cite{CycleGAN2017}, UVCGAN~\cite{UVCGAN2022}, and MUNIT~\cite{MUNIT2018} achieved content-style separation and cycle consistency to facilitate cross-domain attribute transfer without the need for paired data. Despite reducing reliance on labeled data, these unsupervised approaches still face challenges such as incomplete disentanglement, resulting from surrogate objectives and weak inductive biases, leading to cross-attribute interference and insufficient calibration of latent directions, which hampers fine-grained control.

  \section{Methodology}\label{sec:method}
  
  \subsection{Problem Formulation}
\begin{figure}[ht]
      \centering
      \includegraphics[width=0.8\linewidth]{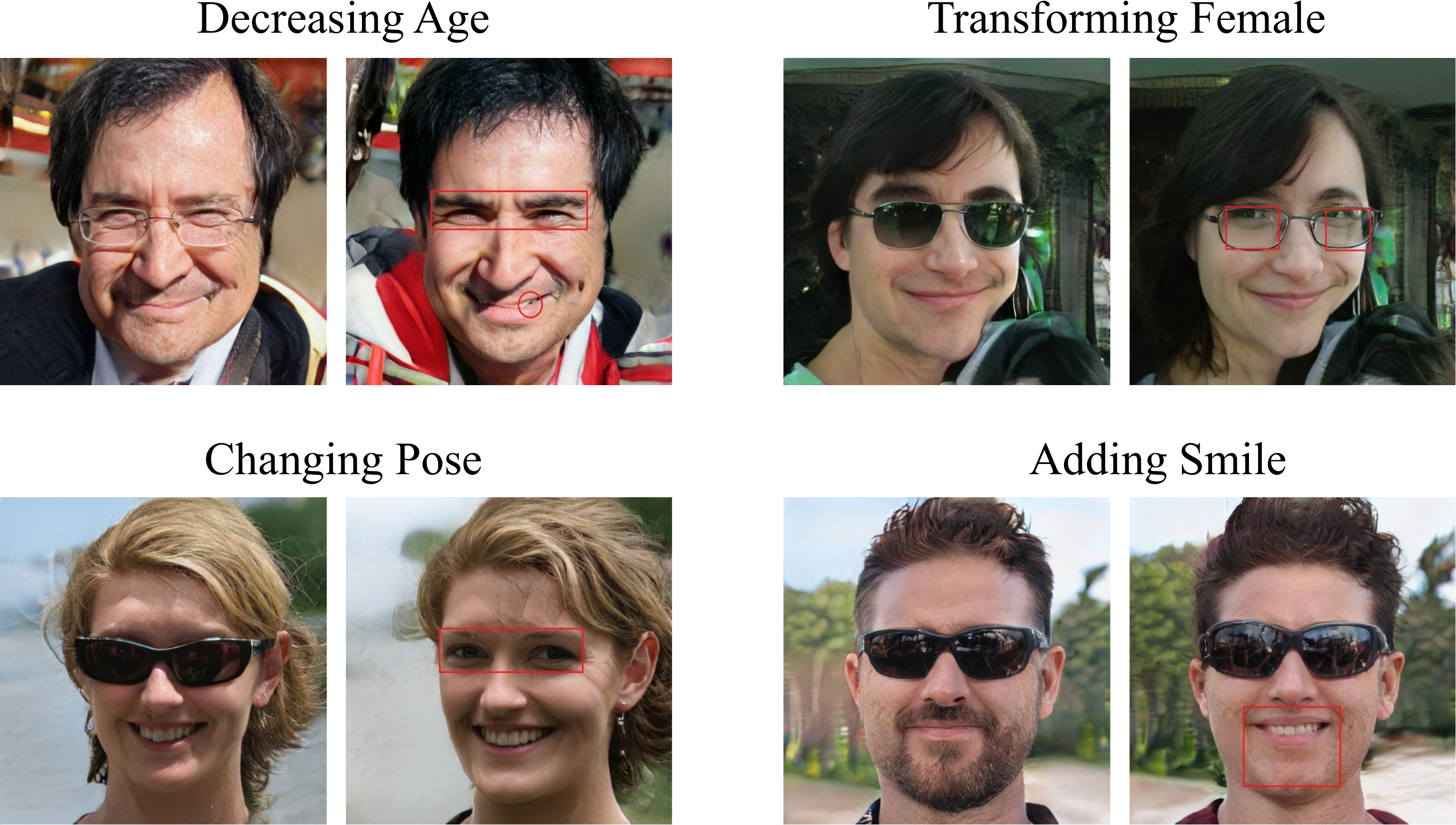}
      \caption{Examples of attribute entanglement. When decreasing age and changing pose, eyeglasses are unintentionally removed; when transforming the face to female, a different pair of eyeglasses is worn; and when adding a smile, the beard is also reduced.}
      \label{fig:sample}
  \end{figure}

\begin{figure*}[ht]
    \centering
    \includegraphics[width=\linewidth]{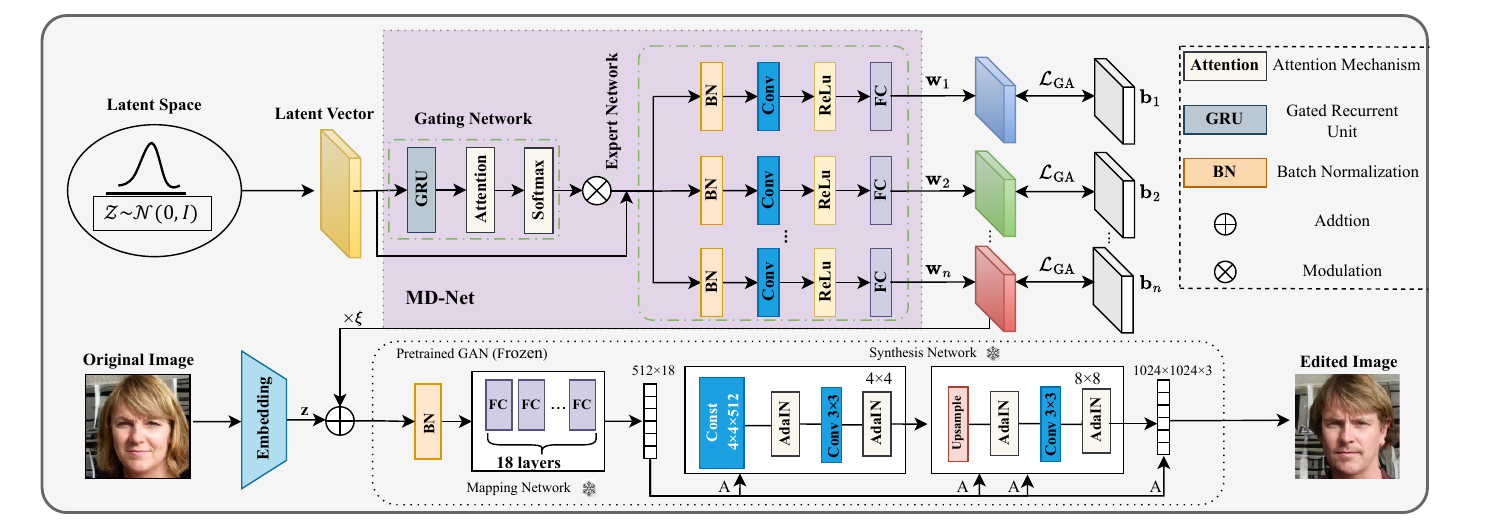}
    \vspace{-18pt}
    \caption{Network architecture of \MyNetwork. It consists of a gating network and an expert network, with the final output being $n$ semantic vectors $\mathbf{w}_1,\mathbf{w}_2,\dots,\mathbf{w}_n$. These semantic vectors are aligned with the corresponding SBVs $\mathbf{b}_1,\mathbf{b}_2,\dots,\mathbf{b}_n$ through a geometry-aware loss $\mathcal{L}_\text{GA}$. The image generator part uses a pre-trained GAN network.}
    \label{fig:arch}
\end{figure*}

  In the GAN's latent space, a \emph{semantic vector} represents a direction for manipulating facial attributes (e.g., smile). Given a pre-trained GAN generator $G$, we sample a latent vector $\mathbf{z} \in \mathbb{R}^{1 \times K}$ and generate a facial image $\mathbf{I}_{ori} = G(\mathbf{z})$. To edit the $i$-th facial attribute, we adjust along its semantic vector $\mathbf{w}_i$: $\mathbf{I}_{edit}=G\!\big(\mathbf{z}+\xi \cdot \mathbf{w}_i\big)$, where $\xi$ is the step size, and $\mathbf{I}_{edit}$ is the edited image. However, semantic vectors are not fully disentangled and attributes are correlated, moving along $\mathbf{w}_i$ may unintentionally alter other attributes, affecting control precision (see Fig.~\ref{fig:sample}). To address this issue and reduce annotation costs, we propose \MyMethod, a label-free disentangled representation learning (DRL) framework designed to learn disentangled semantic vectors for facial attribute editing. Its network architecture, \MyNetwork, is shown in Fig.~\ref{fig:arch}, and the detailed design is introduced below.
  
\subsection{MoE-based Disentangling Network (\MyNetwork)}

To achieve independent semantic representations for each facial attribute and ensure strong attribute disentanglement, we draw inspiration from the Mixture of Experts (MoE)~\cite{dai_2024_deepseekmoe} paradigm. We propose \MyNetwork, a MoE-based disentangling network designed to learn a set of semantic vectors \(\{\mathbf{w}_1, \mathbf{w}_2, \dots, \mathbf{w}_n\}\). \MyNetwork\ comprises a gating network and expert network.

\subsubsection{Gating Network} 
The gating network allocates input data to relevant experts by controlling their activation. We integrate a Gated Recurrent Unit (GRU) to capture the nonlinear relationship between the input and hidden states, providing a memory mechanism that enables the network to make effective decisions based on the current input and historical information, while focusing on salient features through an attention mechanism.

\paragraph{Gated Recurrent Unit (GRU)} 
The GRU captures the nonlinear relationship between the input and the hidden state through its reset gate $\mathbf{r}$ and update gate $\mathbf{u}$:
\begin{align}
\mathbf{r} &= \sigma(\mathbf{W}_r \mathbf{z} + \mathbf{U}_r \mathbf{h}_{init} + \mathbf{b}_r), \\
\mathbf{u} &= \sigma(\mathbf{W}_u \mathbf{z} + \mathbf{U}_u \mathbf{h}_{init} + \mathbf{b}_u),
\end{align}
where \(\sigma(\cdot)\) is the sigmoid function, $\mathbf{z}$ is the input latent vector, $\mathbf{h}_{{init}}$ is the initial hidden state (usually initialized as a zero vector), $\mathbf{W}_r$, $\mathbf{U}_r$, $\mathbf{W}_u$, $\mathbf{U}_u$ are the weight matrices, and $\mathbf{b}_r$ and $\mathbf{b}_u$ are the bias terms. The candidate hidden state is computed by Eq.~\eqref{eq:tilde_h}:
\begin{equation}
  \label{eq:tilde_h}
\tilde{\mathbf{h}} = \tanh \left( \mathbf{W}_h \mathbf{u} + \mathbf{U}_h \left( \mathbf{r} \odot \mathbf{h}_{init} \right) + \mathbf{b}_h \right),
\end{equation}
where \(\odot\) denotes element-wise multiplication,  \(\tanh(\cdot)\) is the hyperbolic tangent function, $\mathbf{W}_h$, $\mathbf{U}_h$ are the weight matrices, and $\mathbf{b}_h$ is the bias terms. The current hidden state then updates via:
\begin{equation}
\mathbf{h} = (1 - \mathbf{u}) \odot \mathbf{h}_{init} + \mathbf{u} \odot \tilde{\mathbf{h}}.
\label{eq:h}
\end{equation}

\paragraph{Attention Mechanism} 
Following the GRU, the attention module maps $\mathbf{h}$ from Eq.~\eqref{eq:h} to $\mathbf{Q}\mathbf{K}\mathbf{V}$:
\begin{align}
\mathbf{Q} &= \mathbf{W}_Q \mathbf{h} + \mathbf{b}_Q, \\
\mathbf{K} &= \mathbf{W}_K \mathbf{h} + \mathbf{b}_K, \\
\mathbf{V} &= \mathbf{W}_V \mathbf{h} + \mathbf{b}_V,
\end{align}
where \(\mathbf{W}_Q, \mathbf{W}_K, \mathbf{W}_V\) weights, and \(\mathbf{b}_Q, \mathbf{b}_K, \mathbf{b}_V\) biases. The attention scores can be computed by Eq.~\eqref{eq:score}:
\begin{equation}
  \label{eq:score}
\mathbf{s} = \frac{\mathbf{Q} \mathbf{K}^\top}{\sqrt{d_k}},
\end{equation}
where \(d_k\) is the dimension of the key vectors. The attention output is a weighted sum:
\begin{equation}
  \label{eq:attention}
\mathbf{a} = \text{softmax}(\mathbf{s}) \odot \mathbf{V},
\end{equation}

\subsubsection{Expert Network}
Facial attribute editing often relies on local feature processing, as different attributes affect specific facial regions, which vary in size and location. For example, gender and expressions are associated with the eye and mouth areas, while age-related changes primarily involve facial contours and skin texture. To address this issue, each expert in the expert network uses tailored convolutional kernels to process different local region features. Specifically, given a latent vector input $\mathbf{z}$, the output of the $i$-th expert is:
\begin{equation}
\label{eq:expert}
E_i(\mathbf{z}) = \text{FC} \left( \text{ReLU} \left( \text{Conv}(\text{BN}(\mathbf{z}), K_i) \right) \right),
\end{equation}
where \(\text{Conv}(\cdot, K_i)\) denotes convolution with kernel \(K_i\) designed for specific local regions, FC and BN represent fully connected layer and batch normalization, respectively. The semantic vector for the \(i\)-th expert combines the gating weight \(a_i \in \mathbf{a}\) from Eq.~\eqref{eq:attention} and the expert output:
\begin{equation}
\mathbf{w}_i = a_i E_i(\mathbf{z}) \in \mathbb{R}^{1 \times K}.
\end{equation}
Finally, the set of semantic vectors $\{\mathbf{w}_i\}$ forms the following semantic vector matrix:
\begin{equation}
\label{eq:concat}
\mathbf{W} = [\mathbf{w}_1, \mathbf{w}_2, \ldots, \mathbf{w}_n]^\top \in \mathbb{R}^{n \times K}.
\end{equation}

\subsection{Loss Function}
\subsubsection{Geometry-aware (GA) Loss}
For each attribute $i$, we use a pre-trained Semantic Boundary Vector (SBV) $\mathbf{b}_i$ to define the decision hyperplane ${H}_i$. Aligning the semantic vector $\mathbf{w}_i$ with $\mathbf{b}_i$ controls the attribute by moving $\mathbf{z}$ along $\pm \mathbf{w}_i$. To prevent cross-attribute interference, $\mathbf{w}_i$ is kept orthogonal to other SBVs. However, due to the nonlinearity of $G$, minimizing $\|\mathbf{w}_i - \mathbf{b}_i\|_2$ alone doesn't guarantee disentanglement, so we introduce a geometry-aware SBV-based loss.\par
Specifically, we first use the Jacobian matrix $\mathbf{J} = \frac{\partial G(\mathbf{z})}{\partial \mathbf{z}} \in \mathbb{R}^{F \times K}$, and let $\mathbf{W} \in \mathbb{R}^{n \times K}$ and $\mathbf{B} \in \mathbb{R}^{n \times K}$ denote the semantic vector and semantic boundary vector (SBV) matrices, respectively. Then, we compute $\mathbf{U} = \mathbf{J}\mathbf{W}^\top$ and $\mathbf{V} = \mathbf{J}\mathbf{B}^\top$, and normalize their columns: $\widehat{\mathbf{U}} = \mathbf{U}\mathbf{D}_\mathbf{U}^{-1}$, $\widehat{\mathbf{V}} = \mathbf{V}\mathbf{D}_\mathbf{V}^{-1}$, where $(\mathbf{D}_{\mathbf{U}})_{ii} = \|\mathbf{U}_{:i}\|_2$ and $(\mathbf{D}_\mathbf{V})_{ii} = \|\mathbf{V}_{:i}\|_2$. Therefore, we define the loss as shown in Eq.~\eqref{eq:j-bv-loss}:
\begin{equation}
    \label{eq:j-bv-loss}
    \mathcal{L}_{\text{GA}} = \big\|\hat{\mathbf{U}}^\top \hat{\mathbf{V}} - \mathbf{I}\big\|_F^2,
\end{equation}
where $\mathbf{I}$ is the identity matrix, $\|\cdot\|_F$ represents the Frobenius norm. This loss enforces alignment with the target direction (($\hat{\mathbf{U}}^\top \hat{\mathbf{V}})_{ii} \to 1$) and suppresses interference from non-target directions ($(\hat{\mathbf{U}}^\top \hat{\mathbf{V}})_{ij} \to 0, i \neq j$).\par
  \subsubsection{Posterior-Prior Alignment (PPA) Loss}
  To stabilize training, we introduce a posterior-prior alignment loss to regularize the latent space and prevent posterior drift. Let $\mathbf{z}$ be a latent vector, and $q_\phi(\mathbf{w}_i \mid \mathbf{z})$ represent the approximate posterior of the $i$-th semantic vector in \MyNetwork. We set the prior as a standard Gaussian, $p(\mathbf{w}i) \sim \mathcal{N}(\mathbf{0},\B{I})$, where $\B{I}$ is the identity matrix. Nonlinear mappings can cause $q_\phi(\mathbf{w}_i \mid \mathbf{z})$ to deviate from the prior, affecting semantic consistency and diversity. To address this, we introduce the following regularization term:
  \begin{equation}\label{eq:kl}
      \begin{split}
      \mathcal{L}_{\text{PPA}} = \frac{\beta}{n} \sum_{i=1}^{n} q_\phi(\mathbf{w}_i \mid \mathbf{z}) \log \frac{q_\phi(\mathbf{w}_i \mid \mathbf{z})^{1/r}}{p(\mathbf{w}_i)^{1/r}},
      \end{split}
  \end{equation}
  where $r > 0$ is the temperature coefficient that controls the softness of the alignment and $\beta > 0$ is the weight of the loss term.\par
  
  Finally, by combining Eq.~\eqref{eq:j-bv-loss}, and~\eqref{eq:kl}, the total loss is given by:
  \begin{equation}\label{total_loss}
      \mathcal{L} = \mathcal{L}_\text{GA} + \mathcal{L}_\text{PPA}.
  \end{equation}
\section{Experiments}\label{sec:experiments}
 \subsection{Experimental Settings}

\textbf{Baselines and Evaluation Metrics.}
We evaluate \MyMethod\ on ProGAN~\cite{ProGAN2018} and StyleGAN~\cite{StyleGAN2019}, comparing it with eight state-of-the-art methods: three supervised (InterFaceGAN~\cite{InterFaceGAN2020}, MDSE~\cite{MDSE2023} and EnjoyGAN~\cite{EnjoyGAN2021}), three unsupervised (label-free) (GANspace~\cite{GANSpace2020}, AdaTrans~\cite{AdaTrans2023}, and SDFlow~\cite{SDFlow2024ICASSP}), and two diffusion-based methods (DDS~\cite{dds2023} and CDS~\cite{cds2024}). All baselines use publicly available models. We measure disentanglement using Attribute Accuracy (AA~\cite{AdaTrans2023}) and Identity Score (IDS~\cite{CA2018}), with higher values indicating better performance. Image quality is evaluated with FID~\cite{NIPS2017FID} and LPIPS~\cite{CVPR2018LPIPS}, where lower FID/LPIPS indicate better results.\par

\textbf{Implementation Details.}
We sample 20,000 latent vectors from the standard Gaussian distribution $\mathcal{N}(0, \B I)$ as training samples for \MyNetwork. Since only four facial attribute edits are implemented, \( n \) is set to 4. The hyperparameters \( \beta \) and \( r \) in Eq. \eqref{eq:kl} are both set to 0.5. During training, the learning rate is set to \( 5 \times 10^{-6} \), and the batch size is set to 2.

  \subsection{Comparative Experiments}
  \label{sec:comparative}
  
  \subsubsection{Quantitative Analysis}
We evaluate \MyMethod\ from two key aspects: disentanglement capability and quality evaluation of edited images.\par
\textbf{Disentanglement Capability.} The results in Table~\ref{tab:disentangled} present AA and IDS scores for ProGAN and StyleGAN. Across all three generators and attributes, \MyMethod\ consistently ranks either first or second. The average AA surpasses 80\% in every case, outperforming the second-best method by more than 5\% on average. Furthermore, \MyMethod\ achieves the highest IDS, indicating its ability to effectively control the target attribute while maintaining identity consistency and preventing undesired alterations to non-target attributes.\par
\begin{table}[ht]
    \centering
    \caption{Quantitative comparison of AA and IDS metrics for various facial attribute editing methods based on the ProGAN and StyleGAN model.}
    \setlength{\tabcolsep}{3pt}
    \renewcommand{\arraystretch}{1}
    \resizebox{0.5\textwidth}{!}{
    \begin{tabular}{l|cccc|cccc}
        \toprule
        \multicolumn{9}{c}{{\large {ProGAN}}} \\ \midrule
        Method & \multicolumn{4}{c|}{AA $\uparrow$ } & \multicolumn{4}{c}{IDS $\uparrow$ } \\ 
        \midrule
        & Age & Gender & Smile & Avg. & Age & Gender & Smile & Avg. \\
        InterFaceGAN~\cite{InterFaceGAN2020} & 0.7498 & 0.7217 & 0.7548 & 0.7421 & 0.7357 & 0.6520 & 0.7513 & 0.7130 \\
        EnjoyGAN~\cite{EnjoyGAN2021} & 0.7532 & \underline{0.7963} & 0.7133 & 0.7543 & 0.7121 & \underline{0.6946} & 0.7647 & 0.7238 \\
        MDSE~\cite{MDSE2023} & 0.7135 & 0.7179 & 0.8758 & 0.7691 & 0.7418 & 0.4270 & \textbf{0.9480} & 0.7056\\ \midrule
        GANspace~\cite{GANSpace2020} & 0.7193 & 0.7138 & 0.8704 & 0.7678 & \underline{0.7820} & 0.6324 & 0.8155 & \underline{0.7433} \\
        AdaTrans~\cite{AdaTrans2023} & 0.7147 & 0.7175 & \textbf{0.8920} & \underline{0.7747} & 0.6694 & 0.6520 & 0.8219 & 0.7144\\
        SDFlow~\cite{SDFlow2024ICASSP} & \underline{0.8354} & 0.7263 & 0.7174 & 0.7597 & 0.7135 & 0.6214 & 0.7855 & 0.7068 \\
        \MyMethod\ (Ours) & \textbf{0.8364} & \textbf{0.8023} & \underline{0.8342} & \textbf{0.8243} & \textbf{0.7933} & \textbf{0.7156} & \underline{0.8754} & \textbf{0.7948}\\
        \midrule
        \multicolumn{9}{c}{{\large {StyleGAN}}}  \\ \midrule
        InterFaceGAN~\cite{InterFaceGAN2020} & \textbf{0.7764} & 0.7147 & \textbf{0.8985} & 0.7965 & 0.6324 & 0.5699 & 0.5803 & 0.5942 \\
        EnjoyGAN~\cite{EnjoyGAN2021} & \underline{0.7431} & 0.7888 & 0.8656 & \underline{0.7992} & \underline{0.7898} & 0.6936 & 0.8219 & 0.7684\\
        MDSE~\cite{MDSE2023} & 0.7417 & 0.7236 & 0.8846 & 0.7833 & 0.6054 & 0.6866 & \underline{0.8746} & 0.7222\\ \midrule
        GANspace~\cite{GANSpace2020} &  0.7056 & 0.7141 & 0.8355 & 0.7517 & 0.5136 & 0.5564 & 0.7431 & 0.6044\\
        AdaTrans~\cite{AdaTrans2023} & 0.7265 & \underline{0.8326} & 0.8386 & 0.7992 & 0.7818 & \underline{0.8096} & 0.8432 & \underline{0.8115}\\
        SDFlow~\cite{SDFlow2024ICASSP} & 0.7198 & 0.7448 & 0.8370 & 0.7672 & 0.7808 &0.6435 & 0.8511 & 0.7585\\
        \MyMethod\ (Ours) & {0.7144} & \textbf{0.8346} & \underline{0.8843} & \textbf{0.8111} & \textbf{0.8345} & \textbf{0.8122} & \textbf{0.8844} & \textbf{0.8437}\\
        \bottomrule
    \end{tabular}
    }
    \label{tab:disentangled}
\end{table}
\begin{table}[ht]
    \centering
    \caption{Quantitative comparison of FID and LPIPS metrics for various facial attribute editing methods based on ProGAN and StyleGAN model (bold numbers indicate the best results, and underlined values denote the second-best results).}
    \setlength{\tabcolsep}{2.5pt}
    \renewcommand{\arraystretch}{1.1}
    \resizebox{0.5\textwidth}{!}{
    \begin{tabular}{l|ccccc|ccccc}
        \toprule 
        \multicolumn{11}{c}{{\large{ProGAN}}} \\ \midrule
        Method & \multicolumn{5}{c|}{FID  $\downarrow$ }  & \multicolumn{5}{c}{LPIPS  $\downarrow$ }\\ \midrule
            & Age & Gender & Pose & Smile & Avg. & Age & Gender & Pose & Smile & Avg.\\
        InterFaceGAN~\cite{InterFaceGAN2020} & 79.10 & 89.52 & 65.06 & \underline{45.82} & 69.88 & 0.3224 & 0.3645 & 0.3933 & \underline{0.2118} & 0.3230 \\  
        EnjoyGAN~\cite{EnjoyGAN2021}  & 78.32 & 77.21 & - & 63.21 & 72.91 & 0.3212 & 0.3254 & - & 0.2936 & 0.3134 \\ 
        MDSE~\cite{MDSE2023} & 77.15 & 77.52 & 66.93 & 74.31 & 73.98 & 0.3487 & 0.4132 & 0.5444 & 0.4097 & 0.4290 \\  \midrule
        GANspace~\cite{GANSpace2020} & \underline{61.46} & \underline{58.15} & \underline{61.96} & {60.88} & \underline{60.61} & {0.3021} & \underline{0.2841} & \textbf{0.2697} & {0.2756} & \underline{0.2829} \\ 
        AdaTrans~\cite{AdaTrans2023}  & 70.65 & 76.47 & - & 75.86 & 74.33 & 0.2938 & 0.3026 & - & 0.3226 & 0.3063 \\ 
        SDFlow~\cite{SDFlow2024ICASSP}  & 62.95 & 74.35 & - & 60.66 & 65.99 & \underline{0.2646} & 0.3149 & - & 0.2329 & 0.2708 \\ 
        \MyMethod\ (Ours) & \textbf{44.36} & \textbf{54.21} & \textbf{49.33} & \textbf{33.46} & \textbf{45.34} & \textbf{0.2233} & \textbf{0.2491} & \underline{0.2991} & \textbf{0.1746} & \textbf{0.2365}\\ 
        \midrule
        \multicolumn{11}{c}{{\large {StyleGAN}}} \\ \midrule
        InterFaceGAN~\cite{InterFaceGAN2020}  &68.93 & 82.38 & 75.65 & \underline{30.65} & 64.40 & 0.3333 & 0.3814 & 0.4112 & \textbf{0.1718} & 0.3244 \\  
        EnjoyGAN~\cite{EnjoyGAN2021} & 74.32 & 76.32 & - &63.39 & 71.34 &0.3856 & 0.3611 & - & 0.3324 & 0.3597  \\ 
        MDSE~\cite{MDSE2023} & 87.54 & 76.99 & 73.11 & 42.36 & 70.00 & 0.4166  & 0.3825 & 0.3548 & 0.2133 & 0.3418 \\  \midrule
        GANspace~\cite{GANSpace2020} & 64.86 & 71.99 & \underline{53.95} & \textbf{25.87} & \underline{54.17} & 0.3119 & 0.3741 & \underline{0.2526} & 0.2133 & 0.2875 \\ 
        AdaTrans~\cite{AdaTrans2023} & 62.34 & 78.59 & -& 68.21 & 67.71 & 0.2644 & 0.3321 & - & 0.3269 & 0.3078 \\ 
        SDFlow~\cite{SDFlow2024ICASSP}  & \underline{59.39} & \underline{68.08} & - & 57.42 & 61.63 & \underline{0.2425} & \underline{0.2641} & - & 0.2235 & \underline{0.2434} \\ 
        \MyMethod\ (Ours) & \textbf{51.23} & \textbf{52.34} & \textbf{49.33} & {41.22} & \textbf{48.53} & \textbf{0.2421} & \textbf{0.2598} & \underline{0.2275} & \underline{0.2021} & \textbf{0.2329} \\ \midrule
    \end{tabular}
    \label{tab:quality}
    }
\end{table}
\begin{table}[!tb]
    \centering
    \caption{A comparison between our proposed \MyMethod\ and state-of-the-art diffusion-based facial attribute editing methods in terms of FID, LPIPS, and inference time (InT).}
    \setlength{\tabcolsep}{3pt}
    \renewcommand{\arraystretch}{1}
    \resizebox{0.5\textwidth}{!}{
    \begin{tabular}{l|cccc|cccc|c}
        \toprule
        Method & \multicolumn{4}{c|}{FID $\downarrow$ } & \multicolumn{4}{c|}{LPIPS $\downarrow$ } & InT $\downarrow$\\ 
        \midrule
        & Age & Gender & Smile & Avg. & Age & Gender & Smile & Avg.\\
        DDS~\cite{dds2023} & 86.34  & 83.57 & 83.36 & 84.42  & \textbf{0.1721} & 0.4122 & 0.2853  & 0.2899  & 17.43s \\
        CDS~\cite{cds2024} & 70.28  & 69.17 & 80.48 & 73.31  & 0.2802 & \underline{0.3024} & \underline{0.2159} & \underline{0.2662} & 16.31s \\
        \MyMethod\ (Ours) & \textbf{51.23} & \textbf{52.34} & \textbf{41.22} & \textbf{48.26} & \underline{0.2421} & \textbf{0.2598} & \textbf{0.2021} & \textbf{0.2347} & \textbf{1.42s} \\

        \bottomrule
    \end{tabular}
    }
    \label{tab:diffusion}
\end{table}
\begin{figure*}[t]
    \centering
    \begin{subfigure}[t]{\textwidth}
        \centering
        \includegraphics[width=0.6\linewidth]{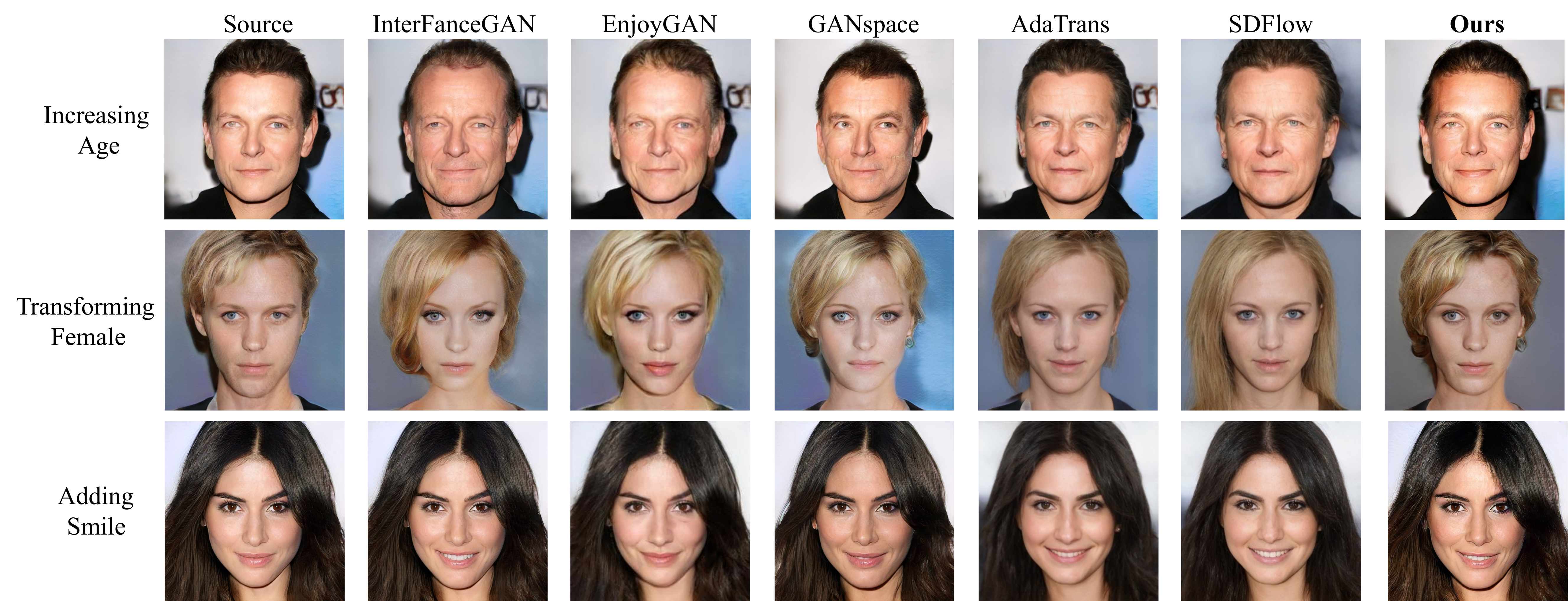}
        \caption{Results on ProGAN: age increase, gender transformation (male $\rightarrow$ female), and smile addition.}
        \vspace{0.3cm}
        \label{fig:progan}
    \end{subfigure}
    \vspace{0.3cm} 
    \begin{subfigure}[t]{\textwidth}
        \centering
        \includegraphics[width=0.6\linewidth]{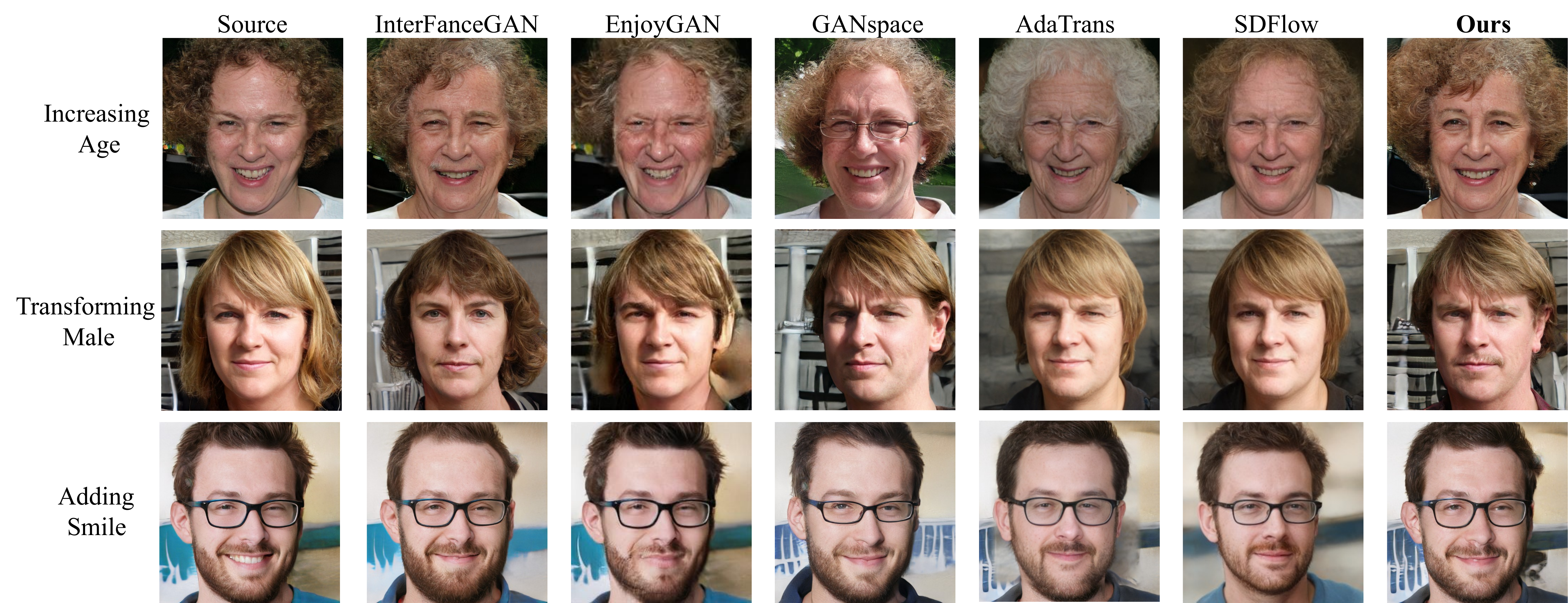}
        \caption{Results on StyleGAN: age increase, gender transformation (female $\rightarrow$ male), and smile addition.}
        \label{fig:stylegan}
    \end{subfigure}
    \vspace{-10pt}
    \caption{Qualitative comparison of facial attribute editing across different generative models. Each subfigure demonstrates three editing operations. The vertical arrangement allows for direct comparison of editing performance across model architectures.}
    \label{fig:gan_comparison_vertical}
\end{figure*}
\begin{figure}[!tb]
    \centering
    \includegraphics[width=0.85\linewidth]{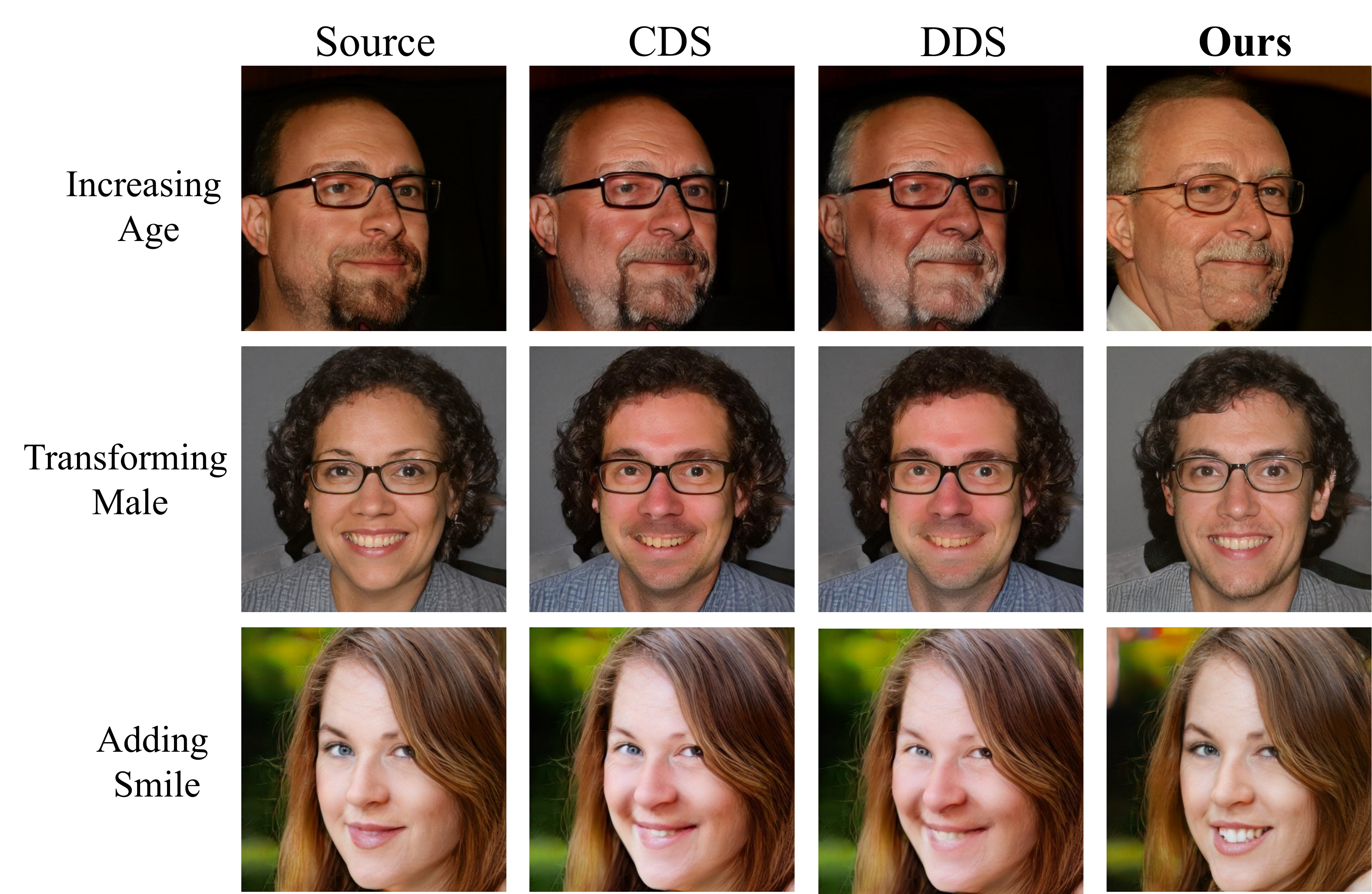}
    \caption{Qualitative evaluation of~\MyMethod\ and diffusion-based baseline methods across the age, gender, and smile attributes.}
    \label{fig:diffuse}
\end{figure}
\textbf{Quality Evaluation of Edited Images.} Table~\ref{tab:quality} reports the performance on ProGAN and StyleGAN. On ProGAN, \MyMethod\ achieves either the best or second-best performance for most attributes, with a 28\% improvement in FID over the second-best method. On StyleGAN, \MyMethod\ outperforms the others across both metrics, surpassing the second-best method by 10\% in FID and 3\% in LPIPS. In Table~\ref{tab:diffusion}, \MyMethod\ is compared to two diffusion-based baselines. It consistently outperforms the alternatives in both FID and LPIPS, while also demonstrating significantly faster inference, producing higher-quality images in less time. These results underscore \MyMethod's superior performance in image quality and inference efficiency across different models.

\subsubsection{Qualitative Analysis}
We evaluate \MyMethod\ and baseline methods through qualitative comparison experiments. Figures~\ref{fig:progan} and \ref{fig:stylegan} show comparisons using ProGAN and StyleGAN, while Figure~\ref{fig:diffuse} illustrates a comparison with two diffusion-based methods. The evaluation focuses on three facial attributes: age, gender, and smile. On ProGAN, both \MyMethod\ and InterFaceGAN produce natural edits, while GANSpace shows limited control, resulting in identity drift. For smile editing, \MyMethod\ maintains facial coherence, while other methods cause distortions. On StyleGAN, \MyMethod\ consistently avoids artifacts and attribute entanglement, preserving both identity and attribute details, whereas EnjoyGAN introduces artifacts in age and gender edits. In the comparison with diffusion-based methods, while they achieve attribute manipulation, they often generate unnatural images with exaggerated lighting or identity distortions. In contrast, \MyMethod\ preserves visual coherence and realism, demonstrating superior overall performance. Overall, \MyMethod\ effectively reduces artifacts, attribute entanglement, and identity drift throughout the editing process.

  \subsection{Ablation Study}
  \label{sec:discussions}
  This section analyzes the contribution of each loss term in~\MyMethod\ and the effect of the temperature coefficient $r$ in the PPA loss. All ablations are conducted on StyleGAN under the same settings as the main experiments.
  
  \subsubsection{Loss Effectiveness Evaluation}
  
\begin{figure}[ht]
    \centering
    \includegraphics[width=0.75\linewidth]{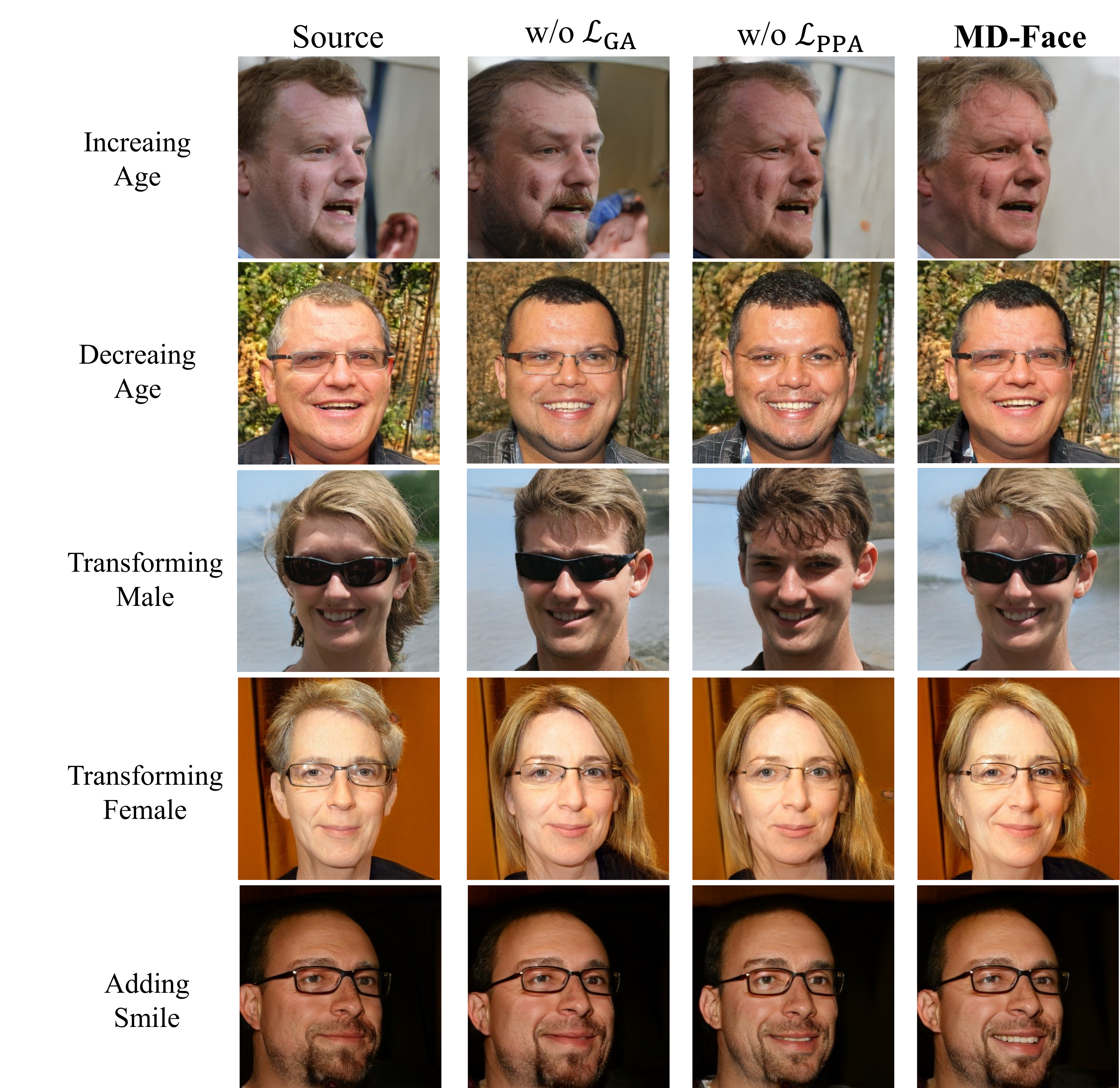}
    \caption{Qualitative ablation results for \MyMethod, comparing the full model with variants without $\mathcal{L}_\text{GA}$ and $\mathcal{L}_\text{PPA}$.}
    \label{fig:Ablation}
  \end{figure}

  To evaluate the impact of each loss component, we remove one loss at a time and report the results in Table~\ref{tab:ablation} and Fig.~\ref{fig:Ablation}. Omitting the posterior-prior alignment loss ($\mathcal{L}_\text{PPA}$) consistently degrades all evaluation metrics, highlighting its importance in preventing overfitting and properly regularizing the semantic vectors. Without the GA loss ($\mathcal{L}_\text{GA}$), attribute edits often lead to mismatches and attribute confusion, significantly reducing AA and IDS, which emphasizes the crucial role of SBV guidance in aligning semantic vectors with target attributes and improving inter-attribute disentanglement.

  \begin{table}[t]
      \centering
      \setlength{\tabcolsep}{5pt}
      \renewcommand{\arraystretch}{0.8}
      \caption{Ablation study quantitative results of \MyMethod. Bold and underlined text indicate the best and second-best results, respectively.}
      \resizebox{0.46\textwidth}{!}{
      \begin{tabular}{l|cccc}
          \toprule
          Method & FID $\downarrow$   & LPIPS  $\downarrow$ & AA  $\uparrow$ & IDS  $\uparrow$\\ \midrule
          \MyMethod\ w/o $\mathcal{L}_{\text{PPA}}$ & 49.22 & 0.2401 & 0.8024 & {0.8024} \\
          \MyMethod\ w/o $\mathcal{L}_{\text{GA}}$ & 55.33 & 0.2587  & 0.6946 & {0.6453} \\ \midrule
          \MyMethod$_{(r=0.1)}$ & \underline{47.23} & 0.2334 & 0.8101 & \underline{0.8433} \\
          \MyMethod$_{(r=0.3)}$ & 47.02 & 0.2344  & 0.8024 & 0.8388 \\
          \MyMethod$_{(r=1)}$ & \textbf{46.54} & 0.2354 & 0.8098 & 0.8377 \\
          \MyMethod$_{(r=3)}$ & 48.12 & \underline{0.2338} & \underline{0.8100} & {0.8425}\\ \midrule
          \MyMethod$_{(r=0.5)}$ (Ours) & {48.53} & \textbf{0.2329}  & \textbf{0.8111} & \textbf{0.8437} \\ \bottomrule
      \end{tabular}
  }
      \label{tab:ablation}
  \end{table}
  
  

  \subsubsection{The Temperature Coefficient \texorpdfstring{$r$}{r} in the PPA Loss}
  \label{sec:temperature_coefficient}
  The 6th-11th rows of Table~\ref{tab:ablation} show that different temperature coefficients $r \in \{0.1, 0.3, 0.5, 1, 3\}$ have only a modest effect on the metrics; in particular, $r=0.5$ consistently yields the best or second-best results, indicating that \MyMethod\ is robust to the choice of $r$.\par
  

  \section{Conclusion}\label{sec:conclusion}

  This paper presents \MyMethod, a novel MoE-based framework for disentangled representation learning (DRL) of facial attributes and controllable manipulation in the GAN latent space. By using MoE to learn semantic vectors and a Jacobian-based pushforward mechanism for aligning semantic boundary vectors (SBVs), \MyMethod\ effectively disentangles facial attributes. Extensive experiments demonstrate that \MyMethod\ outperforms state-of-the-art unsupervised methods in disentanglement and visual quality, while being competitive with supervised and diffusion-based models, and significantly reducing inference latency. However, the current SBV parameterization supports only four attributes, limiting its edit scope. Future work will extend this framework to a broader set of attributes, video, and multimodal facial media, and explore label-free DRL beyond the autoencoder paradigm.

\bibliographystyle{IEEEbib}
\bibliography{D-Face}

\end{document}